\documentclass[10pt,twocolumn,letterpaper]{article}
\pdfoutput=1
\usepackage{cvpr}
\usepackage{times}
\usepackage{epsfig}
\usepackage{graphicx}
\usepackage{amsmath}
\usepackage{amssymb}
\usepackage{mathtools,xparse}

\usepackage{subcaption}

\usepackage{url}
\usepackage{times}
\usepackage[english]{babel}

\usepackage{booktabs} 

\usepackage{algorithm}
\usepackage{algpseudocode}
\usepackage{color}
\usepackage{xcolor}
\usepackage{multirow}

\def\eg{{\cal e.g.,}}
\def\etal{{\em et al.}}
\def\ie{{\em i.e.}}

\newcommand{\btheta}{{\boldsymbol{\theta}}}
\newcommand{\bx}{{\mathbf{x}}}

\newcommand{\bh}{{\mathbf{h}}}
\newcommand{\bbR}{{\mathbb{R}}}

\usepackage[pagebackref=true,breaklinks=true,letterpaper=true,colorlinks,bookmarks=false]{hyperref}
\usepackage[breaklinks=true,bookmarks=false]{hyperref}
 \hypersetup{colorlinks,linkcolor={black},citecolor={black},urlcolor={red}} 
 
 \usepackage{authblk}
 
\author[1,4]{Shih-Yao Lin}
\author[2]{Yen-Yu Lin}
\author[1]{Chu-Song Chen}
\author[3]{Yi-Ping Hung}
\affil[1]{Institute of Information Science, Academia Sinica}
\affil[2]{Research Center for Information Technology Innovation, Academia Sinica}

\affil[3]{Tainan National University of the Arts}
\affil[4]{Tencent America}

\cvprfinalcopy 

\ifcvprfinal\fi
\begin{document}
\title{Learning Conditional Random Fields with Augmented Observations for Partially Observed Action Recognition}



\maketitle

\begin{abstract}
This paper aims at recognizing partially observed human actions in videos. Action videos acquired in uncontrolled environments often contain corrupt frames, which make actions partially observed. Furthermore, these frames can last for arbitrary lengths of time and appear irregularly. They are inconsistent with training data, and degrade the performance of pre-trained action recognition systems. We present an approach to address this issue. For each training and testing action, we divide it into segments, and explore the {\em mutual dependency} between temporal segments. This property states that the similarity of two actions at one segment often implies their similarity at another. We augment each segment with extra alternatives retrieved from training data. The augmentation algorithm is designed in a way where a few alternatives are good enough to replace the original segment where corrupt frames occur. Our approach is developed upon hidden conditional random fields and leverages the flexibility of hidden variables for uncertainty handling. It turns out that our approach integrates {\em corrupt segment detection} and {\em alternative selection} into the process of prediction, and can recognize partially observed actions more accurately. It is evaluated on both fully observed actions and partially observed ones with either synthetic or real corrupt frames. The experimental results manifest its general applicability and superior performance, especially when corrupt frames are present in the action videos.
\end{abstract}

\section{Introduction}
\label{sec:intro}

\begin{figure}
	\centering
	\includegraphics[height=1in]{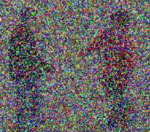}
	\includegraphics[height=1in]{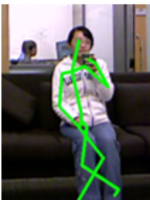}
	\includegraphics[height=1in]{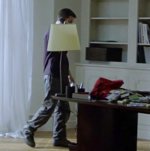}
	\caption{Outlier frames caused by (a) noisy video signals, (b) skeleton inference errors, and  (c) partial occlusions.}
	\label{fig:motivation}
\end{figure}

Video-based human action recognition has been an inherent part in many computer vision applications such as surveillance, robotics, human-computer interaction, and intelligent system. Most research efforts such as \cite{bian12_,zhang17_,liu13learning,ni13_,liu16_transfer,vemulapalli14_,tang15_,yang17_,liu16_,xiaohan15_, lin18_ndt,  lin17_pyramid} focus on recognizing {\em fully observed} human actions in videos. However, the assumption of full observation may not hold in practice due to various issues, including hardware failures, \eg~signal loss or noise~\cite{oshin11_}, software limitations, \eg~skeleton estimation errors~\cite{chaaraoui13_}, and cluttered environments, \eg~partial occlusions~\cite{ayvaci12_,wang09_,lin17_poar}. We consider frames where the above mentioned situations happen {\em outliers}, which make the actions partially observed. Figure~\ref{fig:motivation} shows a few examples of outlier frames caused by diverse issues. In this work, we present an effective approach to recognizing actions with outlier frames.


Outlier frames are inconsistent with training data. They probably cause severe performance degradation of pre-trained recognition systems. A few studies~\cite{cao13_,shen12_,shu12_,wang12_,weinland10_} have attempted to recognize actions with outlier frames. However, they handle outlier frames by using additional domain knowledge and/or assume that outliers are annotated in advance. Thus, their applicability is restricted or extra manual effort is required. Instead, we develop a general approach that infers outlier frames and predicts actions by using the remaining frames, {\em inliers}. Our general approach makes no assumption about the causes of outliers and requires no prior knowledge about the number, locations, and durations of outlier segments. It can work with various features such as those extracted from skeleton structures, RGB, or depth images by conventional or deep learning models.


The task we address is called {\em partially observed action recognition} (POAR), where three major difficulties arise. First, outlier frames need to be identified to exclude their unfavorable effect on recognition. Second, the remaining inliers may carry insufficient information. Third, removing outliers probably makes the action temporally disjointed. Performance gain by applying temporal regularization is not attainable. The approach developed in this work overcomes these difficulties simultaneously. Our idea is that we divide each action into temporal segments, and seek a set of good alternatives to each segment no matter whether this segment is corrupt. A segment is considered corrupt if replacing it with one of its alternatives leads to sufficiently higher confidence in prediction. The substituted alternatives provide extra information and make the action temporally connected, and hence facilitate POAR.

Specifically, every action video is temporally divided into a fixed number of equal-length segments. We carry out {\em alternative augmentation} by leveraging the property of {\em mutual dependency} between segments. This property states that the similarity of two actions at one segment often implies their similarity at another. To augment the $i$th segment of a given action, we use its $j$th segment as the query to the training data, seek the training action with the most similar segment $j$, and retrieve the $i$th segment of that training action as an alternative. Suppose the query, segment $j$, is an inlier, the retrieved alternative is probably of high quality no matter if segment $i$ of that action is an outlier or not. The procedure is repeated for every segment pair. If an action contains a few inlier segments, each of its segments is then augmented with a couple of high-quality alternatives.


After alternative augmentation, we design an approach for training and predicting actions with the extra alternatives. The approach is developed upon {\em hidden-state conditional random fields} (HCRFs)~\cite{quattoni07_}. It leverages hidden variables to model the uncertainty of selecting the original or the alternative observations. With the designed potential functions, our approach can infer outlier segments and seek their alternatives jointly, and hence make a more accurate prediction.


In sum, the main contribution of this work lies in the development of a general approach to partially observed action recognition. It doesn't require any prior knowledge about the number, the durations, and the locations of outlier frames, and can recognize both fully and partially observed actions. Our approach is evaluated
on two datasets, where both synthetic and real outlier frames are present. Compared with several state-of-the-art approaches, our approach demonstrates the effectiveness of outlier frames handling, and achieves remarkably superior results.


\section{Related Work}
\label{sec:relatedwork}

The literature on action recognition is extensive. Our review focuses on approaches that recognize actions in videos.

Due to the recent advances in local descriptors, representing an action in a video as a set of local patches or spatio-temporal cubes, \eg~\cite{laptev05_,maji11_}, is widely adopted for its robustness to possible deformations and occlusions. However, temporal and geometric relationships among local features are ignored, which may lead to suboptimal performance. To address this issue, graphical models such as {\em factorial conditional random fields} (FCRF)~\cite{wang07_} and {\em hidden Markov model} (HMM)~\cite{chen11_} become popular for their expressive power of relationship modeling. Unfortunately, most of these methods recognize only fully observed actions. They are sensitive to outlier frames, and suffer from the performance drop.

Recent approaches adopting features learned by {\em convolutional neural networks} (CNNs)~\cite{krizhevsky12_} have demonstrated their effectiveness in various computer vision applications such as object recognition~\cite{li17_,ShihYF17_}, human pose estimation~\cite{cao16_,chu16_}, tracking~\cite{carneiro13_}, and person re-identification~\cite{xiao16_}. The success of CNNs also sheds light on video-based vision applications. Recent studies of action recognition, \eg~\cite{tran15_,liu16__,donahue15_,liu16__,feichtenhofer16_,gan15_}, focus on using deep learning frameworks for generating more discriminative video representations. Gan~\etal~\cite{gan15_} developed a method that adopts CNNs for high-level video event detection and key-evidence localization. Li~\etal~\cite{li16_} presented a deep network for human action recognition with the aid of multi-granularity information extracted from videos. Simonyan and Zisserman~\cite{simonyan14_} delivered a two-steam ConvNet framework that learns a spatial sub-network and a temporal sub-network at the same time, and achieves very promising performance. However, most of these methods concentrate on recognizing fully-observed actions. They are typically sensitive to outlier frames and suffer from performance degradation when outlier frames are present.

Some research efforts have been made on action recognition with incomplete observation. {\em Early prediction}, \eg~\cite{davis06_,hoai14_,lan14_,raptis13_,ryoo11_,yu12_}, aims to predict an {\em ongoing} action by referring to its beginning part. For instance, Ryoo~\cite{ryoo11_} accomplished this task by using both the {\em integral} and {\em dynamic} bag-of-words. Hoai and De~la~Torre~\cite{hoai14_} developed a max-margin early event detector that identifies the temporal location and duration of an action from the video streaming. On the other hand, Cao~\etal~\cite{cao13_} presented {\em gapfilling} for handling the unobserved frames occurring in an action. They estimated the action likelihood for each observed segment, and inferred the global posterior of the whole action. However, their approach does not take account of temporal coherence between the observed segments. Besides, the approach assumes that the periods of unobserved subsequences have been annotated manually or known in advance. This assumption is less practical in real-world applications.

HCRFs introduce latent variables to model the hidden structures of observations, and have been a powerful model for structured data prediction. Recent studies~\cite{song12_,song13_} have shown that action recognition with HCRFs achieves superior performance to that with HMM and CRFs. However, HCRFs cannot work with incomplete actions. Some studies have attempted to tackle this limitation. Chang~\etal~\cite{chang09_} presented an {\em incremental inference process} to infer HCRFs, and carried out facial expression recognition with incomplete observations. Banerjee and Nevatia~\cite{banerjee14_} proposed a {\em pose filter based HCRF}
(PF-HCRF) model, which uses a detection filter for finding key poses and a root filter for modeling the detected key poses. It infers the temporal locations of the key poses even though the video frames are not fully observed. The methods in~\cite{banerjee14_,chang09_} are able to work with incomplete observations. In this paper, we show a more advanced strategy to deal with incomplete observations: we {\em complete} the observations by borrowing additional segments from training data, and further improve the performance.

In this work, a general approach to partially observed action recognition (POAR) is presented. Regular (fully observed here) action recognition can be considered a special case of POAR if no unobserved part exists. POAR becomes early prediction and gapfilling if there exists merely one unobserved subsequence present at the end and in the middle of the action, respectively. Our method retrieves the alternative segments from training data. It identifies outlier segments, selects their alternatives, and makes the prediction simultaneously. In this manner, our approach bridges the gaps caused by outlier frames, and enriches the required information for making more accurate predictions. Therefore, our approach is general enough to carry out regular action recognition, early prediction, gapfilling. It is also applicable to the cases where training and testing actions are with arbitrary occurrence of outlier frames.

%

\section{The Proposed Approach}

We introduce our approach in this section. A sketch of using HCRFs for action recognition is firstly given. Then, the two key components of the proposed approach, {\em alternative augmentation} and {\em learning HCRFs with augmented observations}, are described, respectively.

\subsection{Action recognition using HCRFs}

A training set of $N$ actions $D =  \{{(\bx_i,y_i)}\}_{i=1}^N$ is given, where each action instance $\bx_i$ is uniformly divided into $T$ temporal segments of the same length, \ie~$\bx_i = \{x_{i,1}, x_{i,2}, ..., x_{i,T}\}$ and $y_i \in {\cal Y}$ is its class. ${\cal Y}$ is the domain of classes. The conditional random fields (CRFs)~\cite{Sutton07} model the conditional probabilities of classes given action instance $\bx$, \ie~$P(y|\bx,\btheta)$, where $\btheta$ is the set of model parameters to be learned. The posterior $P(y|\bx,\btheta)$ in CRFs is a Gibbs distribution, and is written as
\begin{equation}
P(y|\bx,\btheta) = \frac{1}{Z_{\bx}} \exp{(\Psi(y,\bx,\btheta))},
\end{equation}
where $\Psi$ is the {\em potential function}. We will describe it later. $Z_{\bx}$ is the {\em partition function}, which makes $P(y|\bx,\btheta)$ a probability function, namely
\begin{equation}
Z_{\bx}= \sum_{y'\in {\cal Y}}\exp{(\Psi(y',\bx,\btheta))}.
\end{equation}

Parameter set $\btheta$ is derived by {\em maximizing the log likelihood} of the training set $D$:
\begin{equation}
\btheta^* = \arg\max_{\btheta} \sum_{i=1}^N \log{ P(y_i|\bx_i,\btheta) } - \frac{\|\btheta\|^2}{2\sigma^2}, \label{eq:crf_obj}
\end{equation}
where $sigma$ is a positive constant. In Eq.~(\ref{eq:crf_obj}), the first term is the log-likelihood of the training data, and the second one is used for regularization.

Instead of CRFs, we conduct partially observed action recognition on HCRFs, which employ intermediate hidden variables to model the latent structure of observations. The hidden variables whose states are considered {\em key poses} here are used to explore the dependencies among action classes, key poses, and observations as well as to enforce temporal coherence. Specifically, for an action $\bx$, a set of hidden variables $\bh = \{h_1, h_2, ..., h_T\} \in {\cal H}$ is created, one variable for each segment. The conditional probability $P(y|\bx,\btheta)$ in HCRFs is expressed as
\begin{align}
P(y|\bx,\btheta) &= \sum_{\bh \in {\cal H}} P(y, \bh|\bx,\btheta) \label{eq:hcrf_cprob}\\
&= \frac{\sum_{\bh \in {\cal H}}\exp(\Psi(y,\bh,\bx,\btheta))}{\sum_{y' \in {\cal Y},\bh' \in {\cal H}}\exp(\Psi(y',\bh',\bx,\btheta))}. \label{eq:hcrf_cprob2}
\end{align}

\begin{figure}[t]
\centering
\includegraphics[width=0.35\textwidth]{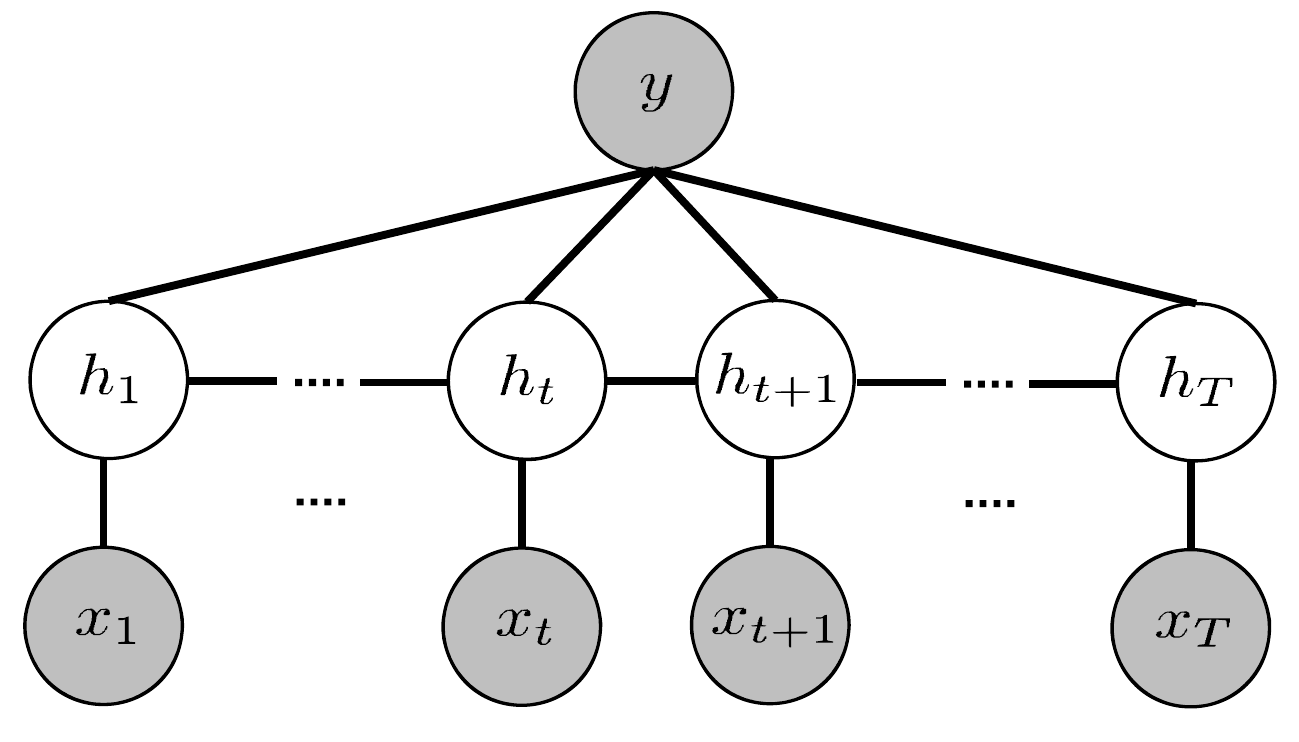}
\caption{The chain-structured HCRFs model.}
\label{fig:chain_}
\end{figure}

Like the original work of HCRFs~\cite{quattoni07_}, we adopt a chain structure shown in Figure~\ref{fig:chain_} to model temporal dependence, and define the potential function as
\begin{align}
 \Psi \left( {y,{\bf{h}},\bx,\btheta } \right) = &\sum\limits_{t=1}^T \langle\phi \left( {{x_t}} \right), {\theta_1}\left( {{h_t}} \right)\rangle  + \sum\limits_{t=1}^T {{\theta _2}\left( {y,{h_t}} \right)}  \nonumber\\
 &\hspace{0.25in}+ \sum\limits_{t=1}^{T-1} {{\theta _3}\left( {y,{h_t},{h_{t+1}}} \right),\label{eq:potential}}
\end{align}
where $\phi(x_t) \in \bbR^d$ is the feature vector of the $t$-th segment of action $\bx$. $\phi(x_t)$ can be any features selected to characterize $x_t$. For instance, we select bag-of-words histograms based on either the {\em cuboid descriptors}~\cite{dollar05_}, $3$D skeleton features, or features learned by deep neural networks in the experiments. $\theta_1(h_t)\in \bbR^d$ is the parameter vector of the $t$-th hidden variable. Inner product $\langle \phi(x_t), \theta_1(h_t)\rangle$ reflects the consensus between observation $x_t$ and hidden state $h_t$. Intuitively, $\theta_1(h_t)$ can be considered as the learned key pose to facilitate action classification. The number of states of each hidden variable $h_t$ corresponds to the number of key poses. $\theta_2(y,h_t)\in \bbR$ measures the compatibility between action class $y$ and hidden state $h_t$. $\theta_3(y,h_t, h_{t+1}) \in \bbR$ represents the consistency between action class $y$ and two successive hidden states $h_t$ and $h_{t+1}$.

Note that our approach can work with the use of general graph structures with various potential functions. We use the chain structure with potential function given in Eq.~(\ref{eq:potential}), because it suffices to get satisfactory results.


With training set $D$ and conditional probability in Eq.~(\ref{eq:hcrf_cprob}), parameter set $\btheta = \{\theta_1,\theta_2,\theta_3\}$ can be optimized by solving  Eq.~(\ref{eq:crf_obj}). Efficient solvers, such as gradient descent based L-BFGS, can be applied to the optimization. After optimization, the HCRFs model $\btheta^*$ is obtained. Given a testing action $\bx$, its label $y$ is then inferred by using loopy belief propagation to solve
\begin{align}
y = \arg\max_{y' \in {\cal Y}} \sum_{\bh \in {\cal H}} P(y',\bh|\bx,\btheta^*). \label{eqn:inference}
\end{align}
Refer to~\cite{quattoni07_} for more details of the training and testing procedures of HCRFs.

\subsection{Alternative Augmentation}
\label{sec:Alternative}

For a corrupt segment $x_t$, the extracted features $\phi(x_t)$ are inconsistent with the learned HCRFs model $\btheta^*$ in potential function Eq.~(\ref{eq:potential}). This issue needs to be handled to avoid substantial performance degradation. The proposed {\em alternative augmentation} aims to augment each segment $x_t$ of every training and testing action $\bx = \{x_t\}_{t=1}^T$ with a set of alternatives no matter if $x_t$ is a corrupt outlier or not. The alternatives are borrowed from training data. Our approach can detect outlier segments and choose proper alternatives to them. It is not necessary that all the alternatives to $x_t$ are of high quality, but just one or few of them are good enough to replace $x_t$ when it is detected as an outlier. In the following, we introduce the proposed alternative augmentation, which is designed based on this requirement.

Alternative augmentation is developed upon the {\em mutual dependency} between segments. Namely, two length-normalized actions are similar at their $j$-th segment. They are likely to be similar at their $t$-th segment. Given the training set $D = \{\bx_i = \{x_{i,t}\}_{t=1}^T\}_{i=1}^N$, we consider an action $\bx = \{x_t\}_{t=1}^T$ to be augmented. To augment the $t$-th segment $x_t$ of $\bx$, we treat its another segment $x_j$ of $\bx$ as the query to $D$, and seek the training action whose $j$-th segment is the most similar to the query. Then, this training action's $t$th segment is employed as an alternative, denoted by $\tilde{x}_{t}^{j}$, \ie
\begin{equation}
\tilde{x}_{t}^{j} \leftarrow x_{i^*,t} \mbox{, where } i^* = \arg\min_{i} \|\phi(x_j) - \phi(x_{i,j})\|.\label{eqn:agumentation}
\end{equation}

\begin{figure}[t]
\centering
\includegraphics[width=3.5in]{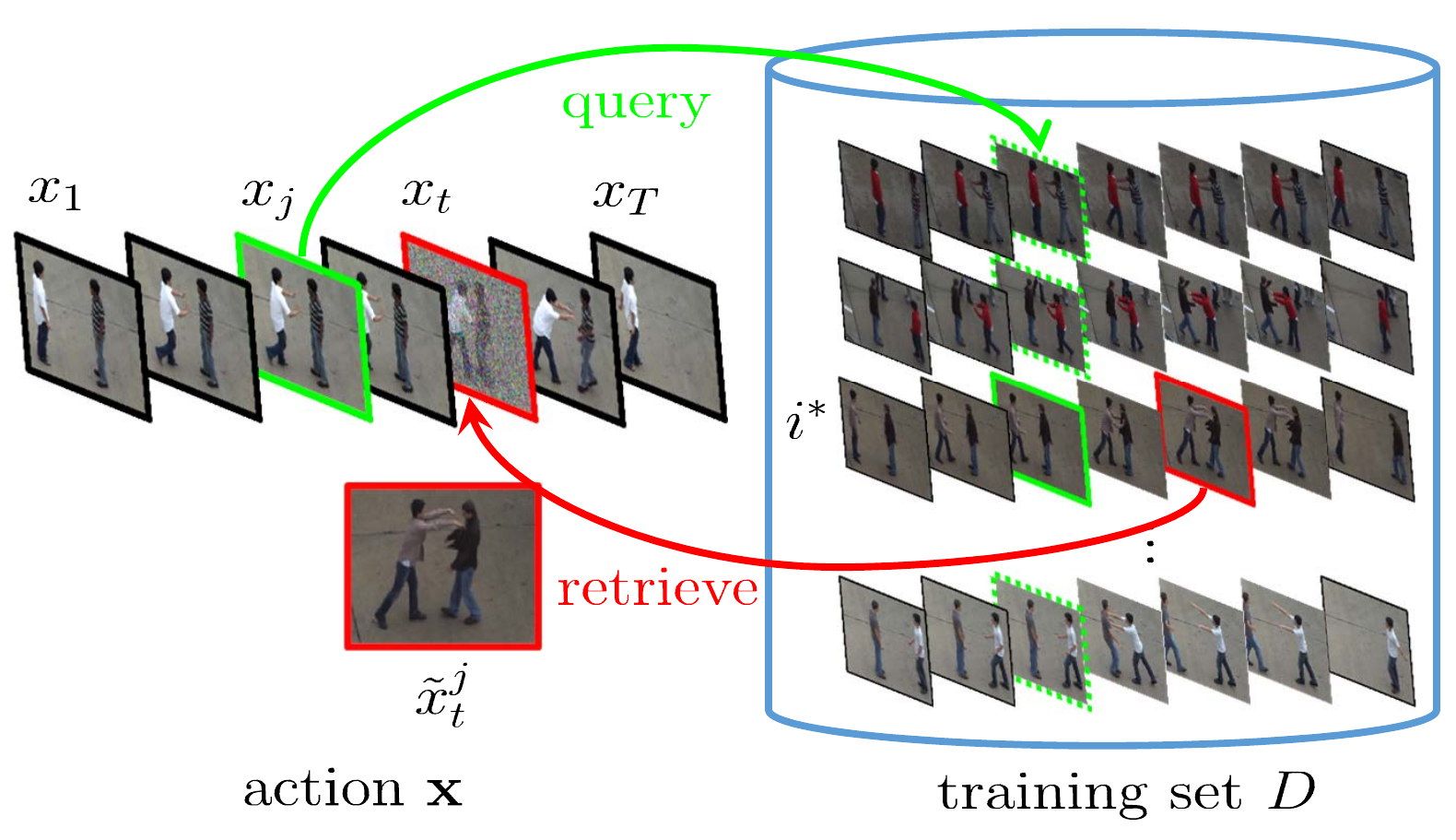}
\caption{Alternative augmentation by mutual recommendation between segments. This figure shows how segment $x_j$ serves as a query to the training set, seeks the training action $i^*$ whose $j$-th segment is the most similar to it, and recommends an alternative $\tilde{x}_t^j$ to another segment $x_t$.}
\label{fig:mainidea}
\end{figure}

For a better understanding, the procedure of mutual recommendation is illustrated in Figure~\ref{fig:mainidea}. By repeating the procedure for every segment pair of $\bx$, the {\em augmented action} of $\bx$, denoted by $\tilde{\bx}$, is yielded where each {\em augmented segment} $\tilde{x}_t$ is composed of the original segment $x_t$ and the $T$ retrieved alternatives $\{\tilde{x}_{t}^{j}\}_{j=1}^T$, \ie
\begin{equation}
\tilde{\bx} = \{\tilde{x}_t\}_{t=1}^T \mbox{, where } \tilde{x}_t =\{x_t,\{\tilde{x}_{t}^{j}\}_{j=1}^T\}.\label{eqn:aug_x}
\end{equation}

Outlier frames may be present in training and testing actions. Therefore alternative augmentation is applied to all training and testing actions. Note that for augmenting a training action, it is tentatively removed from the training set so that all its alternatives come from other training actions. After the procedure, each training or testing action $\bx$ is transformed to the augmented one $\tilde{\bx}$.


Though the augmentation is done in a temporal alignment manner, our approach does not rely on the action videos to be well-aligned frame by frame. It is because in HCRFs, a hidden node would subsume a temporal window of frame-level features, and is tolerant to temporal inconsistency to an extent.  Alternative augmentation can also be extended to be more robust to temporal misalignment between actions via duplicate recommendation. Namely, $\tilde{x}_t^j$ in Eq.~(\ref{eqn:agumentation}) serves as an alterative to not only segment $x_t$ but also its neighboring segments. The main computational cost of augmentation is the {\em nearest neighbor search} (NNS). For an action of $T$ segments, $T^2$ alternatives are found by mutual recommendation. However, NNS is performed $T$ times for augmenting an action of $T$ segments with a careful implementation. Consider Eq.~(\ref{eqn:agumentation}). Once the NNS for segment $x_j$ is finished, the alternatives $\{\tilde{x}_t^j\}_{t=1}^T$ recommended by $x_j$ to the rest segments are known. In addition, algorithms for approximate nearest neighbor search, such as k-d tree or locality sensitive hashing can be applied to further speedup the process.

\subsection{Learning HCRFs with Augmented Actions}

Our approach is designed to work with the augmented actions for POAR. Hence, it needs to detect outliers and select a plausible alternative to each detected outlier. We develop our approach based on HCRFs, where an augmented action $\tilde{\bx} =\{\tilde{x}_t\}_{t=1}^T$ is associated with a set of hidden variables $\tilde{\bh} = \{\tilde{h}_t\}_{t=1}^T$, one hidden variable $\tilde{h}_t$ for each augmented segment $\tilde{x}_t = \{x_t, \{\tilde{x}_{t}^{j}\}_{j=1}^T\}$.

We leverage the hidden variables in HCRFs to model the uncertainty about both {\em poses} and {\em observations}. Specifically, the hidden variable here is {\em composite}, \ie~$\tilde{h}_t = [h_t^o \in \{0, 1, ..., T\}, h_t^p \in \{1, 2, ..., S\}]$, where $T$ and $S$ are the numbers of the alternatives and latent poses, respectively. Element $h_t^o$ specifies which {\em o}bservation is picked at time stamp $t$. It takes value $0$ if the original segment (observation) $x_t$ is identified as an inlier and picked. When $h_t^o$ takes value $j \in [1,T]$, $x_t$ is detected as an outlier and replaced by its $j$-th alternative $\tilde{x}_{t}^{j}$. Element $h_t^p$, like the hidden variables used previously, corresponds to the latent poses.

\begin{figure}[t]
	\centering
	\includegraphics[width=0.48\textwidth]{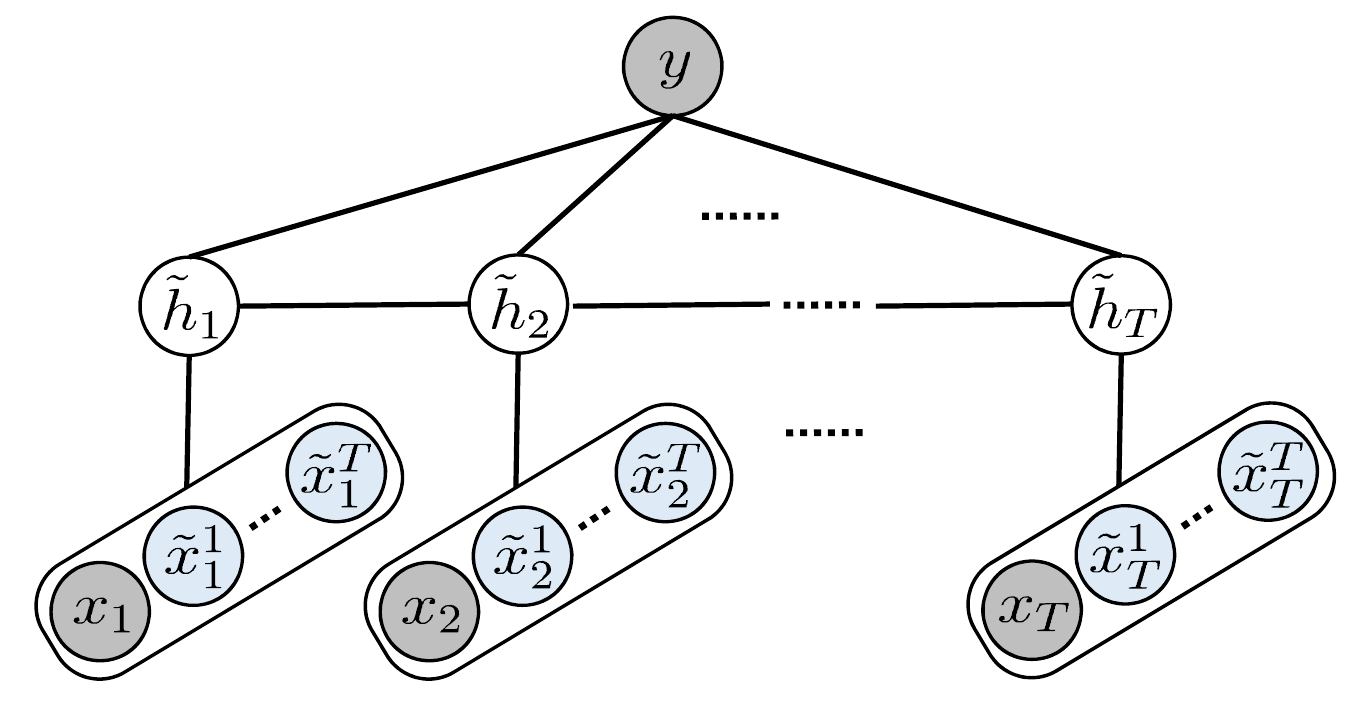}
	\caption{Our extended HCRFs model for working on actions with augmented observations.}
	\label{fig:oahcrf_model}
\end{figure}

The chain-structured model of our approach is shown in Figure~\ref{fig:oahcrf_model}.
Compared to that in Figure~\ref{fig:chain_}, each hidden variable is composite, and each observation node contains not only the original segment but also the $T$ alternatives. To work with the composite variables and augmented observations, the potential function is generalized from Eq~(\ref{eq:potential}), and is defined as follows
\begin{align}
\Psi \left( {y,{\bf{h}},\tilde{\bx},\btheta } \right) &= \sum\limits_{t=1}^T \langle\phi ( {{\tilde{x}_t}} ), {\theta_1}( {{\tilde{h}_t}} )\rangle + f(\tilde{h}_t) \nonumber\\
&\hspace{-0.2in}+ \sum\limits_{t=1}^T {{\theta_2}( {y,{\tilde{h}_t}} )} + \sum\limits_{t=1}^{T-1} {{\theta_3}( {y,{\tilde{h}_t},{\tilde{h}_{t+1}}}),\label{eq:aug_potential}}
\end{align}
where $\phi(\tilde{x}_{t})$ is the feature representation of augmented segment $\tilde{x}_{t}$, and is defined as the concatenated column vector of all its $(1+T)$ elements,
\begin{equation}
\small{\phi(\tilde{x}_{t}) = \left[  \begin{array}{c} \phi^{\top}(x_t) \;\; \phi^{\top}(\tilde{x}_t^1) \;~\cdots~\; \phi^{\top}(\tilde{x}_t^T) \end{array} \right]^{\top} \in \bbR^{(1+T) \cdot d}}. \label{eqn:aug_phi}
\end{equation}
The parameter vector $\theta_1(\tilde{h}_t)$ corresponding to composite variable $\tilde{h}_t$ takes both the picked observation and pose into account, and is expressed as
\begin{equation}
\small{\theta_1(\tilde{h}_t) = \theta_1(h_t^o = j, h_t^p = k) = \left[  \begin{array}{l} \mathbf{0~~} \in \bbR^{j \cdot d} \\ \mathbf{\lambda}_k \in \bbR^{d} \\ \mathbf{0~~} \in \bbR^{(T-j) \cdot d} \end{array} \right],}\label{eqn:aug_theta1}
\end{equation}
where $\mathbf{\lambda}_k$ is the parameter set of the $k$-th pose to be learned, and $\mathbf{0}$ is a vector whose elements are $0$. Function $f(\tilde{h}_t)$ is used to express our bias towards picking the original segment, and is given by
\begin{equation}
f(\tilde{h}_t) =
\begin{cases}
\epsilon,& \text{if $h_t^o = 0$,}   \\%
0,& \text{otherwise,}
\end{cases}
\label{eqn:fun_f}
\end{equation}
where $\epsilon$ is a non-negative constant.

The first term corresponding to composite hidden variable ${\tilde{h}_t}=[h_t^o = j, h_t^p = k]$ in Eq.~(\ref{eq:aug_potential}) becomes
\begin{equation}
\langle\phi ( {\tilde{x}_t} ), \theta_1(\tilde{h}_t)\rangle + f(\tilde{h}_t)  =
\begin{cases}
\langle\phi(x_t), \mathbf{\lambda}_k\rangle+\epsilon,& \text{if $j = 0$,}   \\%
\langle\phi(\tilde{x}_t^j), \mathbf{\lambda}_k\rangle,& \text{otherwise.}
\end{cases}
\label{eqn:sigular_term}
\end{equation}
It measures the compatibility between the $k$th latent pose and the $j$th alternative (or the original segment if $j=0$). If the original segment is picked, extra value $\epsilon$ is added. It ensures that the original segment is replaced only when the substituted alternative is sufficiently better. The other two terms in Eq.~(\ref{eq:aug_potential}), $\theta_2(y,{\tilde{h}_t})\in \bbR$ and $\theta_3(y,\tilde{h}_t,\tilde{h}_{t+1})\in \bbR$, evaluate the consistence among adjacent hidden variables and the class label. We simply set ${\theta_2}( {y,{\tilde{h}_t}} ) = \theta_2(y,h_t^p) \in \bbR$ and $\theta_3( y,\tilde{h}_t,\tilde{h}_{t+1}) = \theta_3( y,h_t^p,h_{t+1}^p)\in \bbR$. Namely, they are the same as those in Eq.~(\ref{eq:potential}).

Composite hidden variable $\tilde{h}_t = [h_t^o \in \{0, ..., T\}, h_t^p \in \{1, ..., S\}]$ can be converted into a single one with $(1+T)S$ states. With the new potential in Eq.~(\ref{eq:aug_potential}), HCRFs model $\btheta^*$ can be learned with augmented actions by optimizing Eq.~(\ref{eq:crf_obj}). The learned model then predicts novel augmented actions via Eq.~(\ref{eqn:inference}).

The conditional probability in HCRFs in Eq.~(\ref{eq:hcrf_cprob}) is inferred by summing all configurations of hidden variables. A configuration of hidden variables specifies how the original segment or one of its alternatives is picked at each time stamp of an augment action. In Eq.~(\ref{eq:hcrf_cprob}), the conditional probability is computed by taking the exponential of the potentials of all configurations, so it is dominated by the configuration with the maximal potential value. From the inferred configuration with the maximal potential, it can be realized that our approach recognizes a partially observed action by detecting outliers and picking their alternatives.

As reported in the paper of HCRFs~\cite{quattoni07_}, the key step of HCRFs, belief propagation, is of complexity ${\cal O}(|{\cal Y}|T|{\cal H}|^2)$, where $|{\cal Y}|$, $T$, and $|{\cal H}|$ are the numbers of classes, segments, and hidden states, respectively. In addition, the nearest neighbor search is performed $T$
times for augmenting an action. On UT-Interaction $\#1$ with $|{\cal Y}| = 6$, $T=20$, and $|{\cal H}|=252$, the average running time of augmenting an action and predicting it is $1.98$ seconds on a modern PC with an Intel $i7-4770$~$3.40$GHz processor using {\tt C++} implementation.

\begin{figure*}
	\begin{center}
		\begin{tabular}{cccccccc}
			\hspace{-0.1in}
			\includegraphics[height=1.1in]{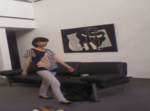}&
			\hspace{-0.15in}
			\includegraphics[height=1.1in]{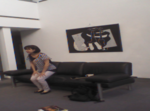}&
			\hspace{-0.15in}
			\includegraphics[height=1.1in]{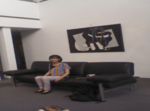}&
			\hspace{-0.15in}
			\includegraphics[height=1.1in]{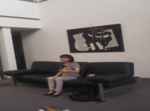}&
			\hspace{-0.15in}
			\includegraphics[height=1.1in]{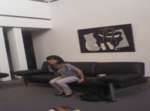}&
			\hspace{-0.15in}
			\includegraphics[height=1.1in]{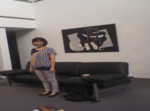}&
			\hspace{-0.15in}
			\includegraphics[height=1.1in]{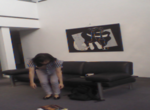}&
			\hspace{-0.15in}
			\includegraphics[height=1.1in]{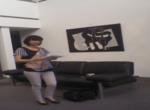}\\
			\hspace{-0.15in}
			{\small   Walk}&
			\hspace{-0.15in}
			{\small  Sit down}&
			\hspace{-0.15in}
			{\small  Sit still}&
			\hspace{-0.15in}
			{\small  Use a TV remote}&
			\hspace{-0.15in}
			{\small  Stand up}&
            \hspace{-0.15in}
            {\small  Stand still}&
			\hspace{-0.15in}
			{\small  Pick up books}&
			\hspace{-0.15in}
			{\small  Carry books}\\
			\hspace{-0.15in}
			\includegraphics[height=1.1in]{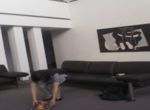}&
			\hspace{-0.15in}
			\includegraphics[height=1.1in]{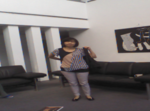}&
			\hspace{-0.15in}
			\includegraphics[height=1.1in]{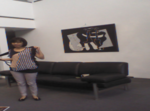}&
			\hspace{-0.15in}
			\includegraphics[height=1.1in]{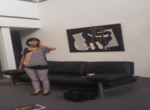}&
			\hspace{-0.15in}
			\includegraphics[height=1.1in]{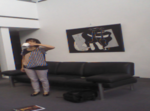}&
			\hspace{-0.15in}
			\includegraphics[height=1.1in]{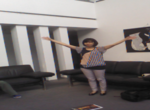}&
			\hspace{-0.15in}
			\includegraphics[height=1.1in]{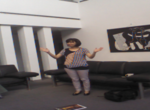}&
			\hspace{-0.15in}
            \\
            \hspace{-0.15in}
			{\small  Put down books}&
			\hspace{-0.15in}
			{\small  Carry a backpack}&
			\hspace{-0.15in}
			{\small Drop a backpack}&
			\hspace{-0.15in}
			{\small Make a phone call}&
			\hspace{-0.15in}
			{\small Drink water}&
			\hspace{-0.15in}
			{\small Wave hand}&
			\hspace{-0.15in}
			{\small Clap}&
            \hspace{-0.15in}
		\end{tabular}
	\end{center}
	\caption{The daily activities $3$D dataset collected us. One example comes from each of the fifteen action categories.}
	\label{fig:CITI_dataset}
\end{figure*}

\section{Experimental Setup}
\label{exp_setup}


In this section, we describe the settings of the conducted experiments, including two datasets used for performance evaluation, the adopted feature representations, and the evaluation metrics on each of the two datasets.

\subsection{Datasets for Performance Evaluation}
\label{subsec:database}

Our approach is evaluated on a daily activities dataset we collected, {\em CITI-DailyActivities3D}\footnote{CITI-DailyActivities3D dataset is available at \url{https://sites.google.com/view/citi3ddataset/}} and a benchmark dataset, {\em UT-Interaction}~\cite{Ryoo10database}.
The first one contains actions with outlier frames occurring irregularly and naturally. The second one consists of {\em clean} actions. Thus, synthetic outlier frames are added. The primary goal of evaluation on the first dataset is to measure how our approach performs in realistic cases. The goal on the second one is to analyze how well it performs when different fractions of outlier frames are present. The two datasets contain videos of different modalities, such as RGB videos and $3$D skeleton structures, and cover actions ranging from single-person actions and multi-people interactions.

\subsubsection{CITI-DailyActivities3D dataset}
\label{subsec:citi_description}

This work delivers an integrated solution to outlier detection, alternative selection, and action prediction. Existing benchmarks of action recognition, \eg~\cite{song13_,oreifej13_,xia13_}, contain videos where no or few corrupt frames show. For a more realistic evaluation, we adopt this dataset where outlier frames are present.

Ten actors were employed to perform fifteen daily activities in the construction of this dataset. One of the ten actors is left-handed. The fifteen daily activities involve {\em walk, sit down, sit still, use a TV remote, stand up, stand still, pick up books, carry books, put down books, carry a backpack, drop a backpack, make a phone call, drink water, wave hand, and clap}. Figure~\ref{fig:CITI_dataset} displays one example from each of the fifteen categories. Microsoft Kinect is used in the collection so that the RGB videos and the depth maps are available simultaneously. The skeleton streams are also attainable by applying the method in~\cite{shotton13_} to the depth maps.

\begin{figure}[t]
	\centering

		\includegraphics[height=0.75in]{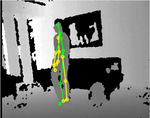}
		\label{fig:occlusion1}
		\includegraphics[height=0.75in]{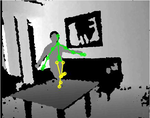}
		\label{fig:occlusion2}
		\includegraphics[height=0.75in]{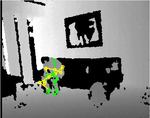}
		\label{fig:occlusion3}

	\caption{Outliers in skeleton streams caused by (a) self-occlusion, (b) occlusions by other objects, and (c) skeleton inference errors.}
	\label{fig:occlusion}
\end{figure}

The resultant dataset is challenging. Outlier frames caused by different issues can occur at any temporal positions with arbitrary durations in videos. The dataset is composed of $482$ skeleton sequences. Among them, $300$ sequences are clean. More than $10\%$ of frames in each of the other $182$ sequences are outliers. Some outlier frames in the skeleton streams are shown in Figure~\ref{fig:occlusion}. The part in yellow represents the extracted skeletons with low confidence. Similar to many existing benchmarks, difficulties such as large intra-class variations, high inter-class similarity, and different perspective settings, present in this dataset.


\subsubsection{UT-Interaction dataset}
\label{subsec:ut_description}

This database collects high-level human interaction videos of six activity categories, including {\em hand-shaking, hugging, kicking, pointing, punching,} and {\em pushing}. The dataset has two subsets, \ie~UT-Interaction $\#1$ and $\#2$. Each subset contains $60$ videos of the six types of human interactions. Both segmented and unsegmented versions of this dataset are available. Like approaches for comparison, we choose the former for evaluation.

\begin{figure}[]
	\centering
	\includegraphics[height=0.68in]{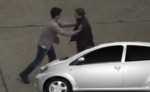}
	\includegraphics[height=0.68in]{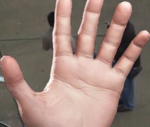}
	\includegraphics[height=0.68in]{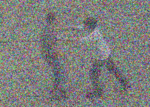}
	\caption{Outlier frames caused by (a) \& (b) occlusions by other objects and (c) signal noise. The average ratios of the noisy area to the whole image are $34.5$\%, $53.7$\% and $100$\% in the three cases.}
	\vspace{-0.1in}
	\label{fig:OT_db}
\end{figure}

We added artificial outlier frames to the videos for evaluation. A wide range of {\em outlier ratio}, \ie~outlier frames to all frames, from $0$ to $0.8$ is considered. The types of the artificial outlier frames include signal noise and occlusions by various objects. Some examples of these synthetic outliers are shown in Figure~\ref{fig:OT_db}.


\subsection{Feature Representation and Evaluation Metrics}

We represent actions in our CITI-DailyActivities3D dataset based on the absolute $3$D body joint positions in the skeleton streams. Each action is uniformly sampled $T=30$ skeletons. To make the representation more robust, we first transform from the world coordinate system to the person-centric coordinate for the skeletal data by setting the hip center at the origin. Then, a skeleton in this dataset is randomly chosen as the reference. All the other skeletons are normalized so that their body part lengths can be the same as that of the reference. Finally, we rotate each skeleton so that the ground plane projection of the vector from its left hip to its right hip is parallel to the global $x$-axis.

For UT-Interaction dataset, we follow the method in \cite{ryoo11_}, where the {\em spatial-temporal interest points} (STIPs) are firstly detected. Then, the \emph{cuboid descriptor}~\cite{dollar05_} is applied to each of the detected STIPs. STIPs are detected by using the {\em Harris$3$D corner detector}~\cite{laptev05_} in this work. Then, actions are represented by using the {\em bag-of-words model} \cite{fei05_}, where the {\em visual words} are generated via the $k$-means clustering algorithm with $800$ centers. Each training and testing action is partitioned into $T=20$ equal-length segments. A bag-of-words histogram is compiled for each segment. For this dataset, features learned by deep neural networks are also adopted.

Unless further specified, our approach and all the competing approaches use the same feature representation on each dataset in the experiments for fair comparison. In our CITI-DailyActivities3D dataset, we split the ten subjects into two equal-size groups. The actions from one group firstly serve as the training data, while the rest as the testing data. We then switch the two subject groups. The average performance is reported. For the UT-Interaction dataset, we follow~\cite{banerjee14_,cao13_,ryoo11_}, and use {\em leave-one-sequence out cross validation} for evaluating the performance.


\begin{table*}[]
	\caption{Accuracy rate ($\%$) of various approaches on the dataset we collected.}
	\vspace{-0.1in}
	\label{table:citti_results}
	\small{
		\begin{center}
			\begin{tabular}{lccc}
				\toprule
				Method                                                &    Task $\#1$ \quad &   Task $\#2$  \quad  &    Task $\#3$\\
				\midrule
				na\"{\i}ve Bayes classifier  (NBC)                                    & $73.3$ & 64.8 & 69.2\\
				recurrent neural networks (RNNs)~\cite{martens11_}                   & $77.3$ & 68.1 & 71.7 \\
				hidden Markov model (HMM)                                                          & $73.3$ & 51.6 & 64.2 \\
				hidden conditional random fields (HCRFs)~\cite{quattoni07_}                & $80.3$ & 60.3 & 68.8  \\
				hierarchical sequence summarization (HSS)~\cite{song13_}            & \bf{84.0} & 61.5 & 66.3  \\
				approach by Gowayyed~\etal~\cite{gowayyed13_}          & $83.0$ & 68.3 & 74.6 \\
				\midrule
				Ours                                                  &80.0 & \bf{74.1}& \bf{79.8} \\
				\bottomrule \end{tabular}
	\end{center}}
	\vspace{-0.15in}
\end{table*}

\section{Experimental Results}

In this section, our approach is evaluated on the dataset we collected and the UT-Interaction dataset. We report and the results.


\subsection{Results on CITI-DailyActivities3D dataset}
\label{sec:realsitic_database}

Three evaluation tasks are conducted on this dataset. Task~$\#1$ aims at evaluating the performance of approaches on {\em fully-observed} videos. Namely, both training and testing action contain no outlier frames. Task~$\#2$ and Task~$\#3$ focus on the tolerance of approaches to outliers. In Task~$\#2$, approaches are learned with clean training data, but are tested on actions with outliers. In Task~$\#3$, both the training and testing sets are the mixtures of clean and corrupt actions. In Task~$\#1$, we check if our approach with extra components for outlier handling still performs well on clean actions. More importantly, we are interested in the performance gaps between the first task and the other two tasks, which reveal the robustness of an approach to outliers.

We select six existing approaches for comparison, including {\em Na\"{\i}ve Bayes classifier} (NBC), {\em recurrent neural networks} (RNNs)~\cite{martens11_}, {\em hidden Markov model} (HMM), {\em hidden-CRFs} (HCRFs)~\cite{quattoni07_}, {\em hierarchical sequence summarization} (HSS) model~\cite{song13_}, and the approach by Gowayye~{\em et al.}~\cite{gowayyed13_}.

We particularly focus on the comparison between HCRFs and ours. Both methods are established on HCRFs and use the same inference algorithms for training and testing. Two main technical components, alternative augmentation and the extended model for working on augmented actions, distinguish our approach from HCRFs for outlier handling.

Except that in~\cite{gowayyed13_}, all the approaches adopt the $3$D skeleton features that we compiled. For graphical model-based classifiers, \eg~HMM, HCRFs, and ours, the feature vector at each segmentation is the representation of the corresponding observation node. For classifiers working on data with representations considering the whole videos, \eg~NBC, we concatenate the feature vectors of all frames. The approach by Gowayye~\etal~\cite{gowayyed13_} takes into account the features based on body joint trajectories and uses {\em Fourier temporal pyramid} (FTP). The recognition rates of all approaches on the three tasks are reported in Table~\ref{table:citti_results}.

\textbf{Results on Task $\#1$.} The baseline NBC gives the accuracy rate of $73.3\%$. The graphical model-based approaches, including RNN, HMM, HCRFs, and HSS, achieve the accuracy between $73.3\%$ and $84.0\%$. The method of Gowayyed \etal~\cite{gowayyed13_} reaches $83.0\%$. Our approach gets the recognition rate of $80.3\%$. It is comparable to most competing approaches. Note that HCRFs and our approach give almost the same recognition rates in this task with the {\em clean} data. It means that the additional mechanisms of our approach do not cause performance drop, even though they are designed to handle outliers.


\textbf{Results on Task $\#2$.} The major difference between Task~$\#1$ and Task~$\#2$ is that the testing actions in the latter contain outliers. Compared the performance on the two tasks, all the six competing approaches suffer from substantial performance drops ranging from $8.5\%$ ($=73.3\% - 64.8\%$ in NBC) to $22.5\%$ ($=84.0\% - 61.5\%$ in HSS). We also observe that the drops are even more dramatic in graphical model-based approaches, such as HMM and HCRFs, since their complex models are more sensitive to noisy data. The features and the FTP structure used in~\cite{gowayyed13_} show their robustness on this task. Our approach is designed to address outliers. It detects outliers, and replace them with plausible alternatives. It turns out that the drop is only $5.9\%$ ($=80.0\% - 74.1\%$), even if our approach is established upon graphical models. The achieved accuracy $74.1\%$ is more favorable than those by all other approaches.

\textbf{Results on Task $\#3$.}

The difference between this task and the two previous ones is that the training actions also contain outlier frames. Comparing the accuracy in Task~$\#1$ and Task~$\#3$, all the six competing approaches still suffer from severe performance degradation. The outlier frame distributions in training and testing data are similar. Task~$\#3$ may not be more difficult than Task~$\#2$ though both training and testing actions have outlier frames in Task~$\#3$. To sum up, the results indicate that our approach can work with not only corrupt testing data but also corrupt training data, and outperform the competing approaches significantly.

In order to evaluate the sensitivity of our method to the number of hidden states in HCRFs, we conduct an experiment to quantify the parameter sensitivity. The performance of our approach with different numbers of hidden states regarding key poses on all the three tasks are shown in Figure~\ref{fig:hiddenpose}. The results point out that a few hidden states, \eg~$20$, suffice for getting the stable performance in all the three tasks. Actually except those in Figure~\ref{fig:hiddenpose}, the performances of our approach are reported in all the experiments by setting the number of hidden states regarding key poses to $20$.

\begin{figure*}[t]
	\begin{center}
			\hspace{-0.1in}
			\includegraphics[height=1.4in]{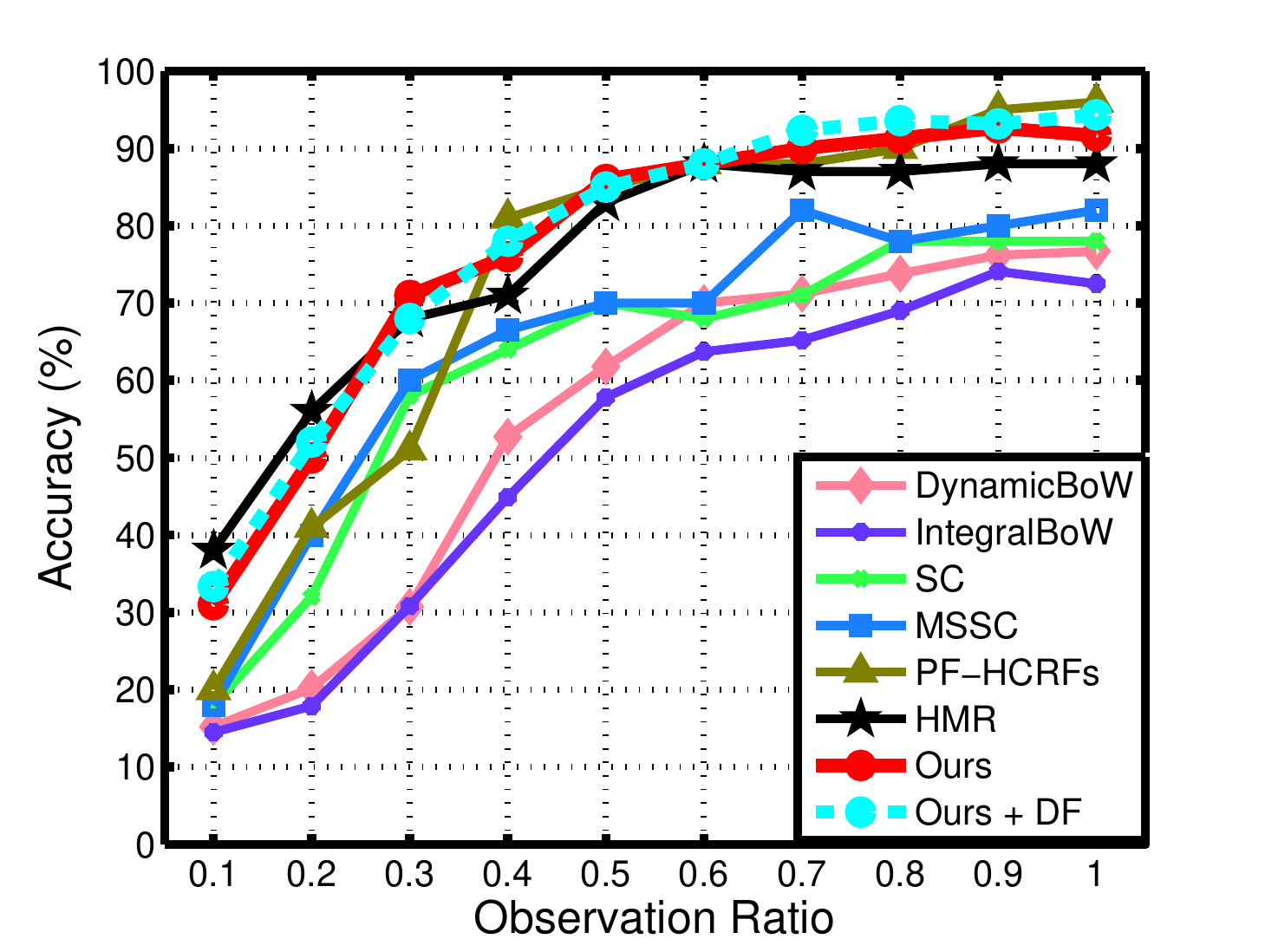}
			\label{fig:UT1_Prediction}\hspace{-0.25in}			
			\includegraphics[height=1.4in]{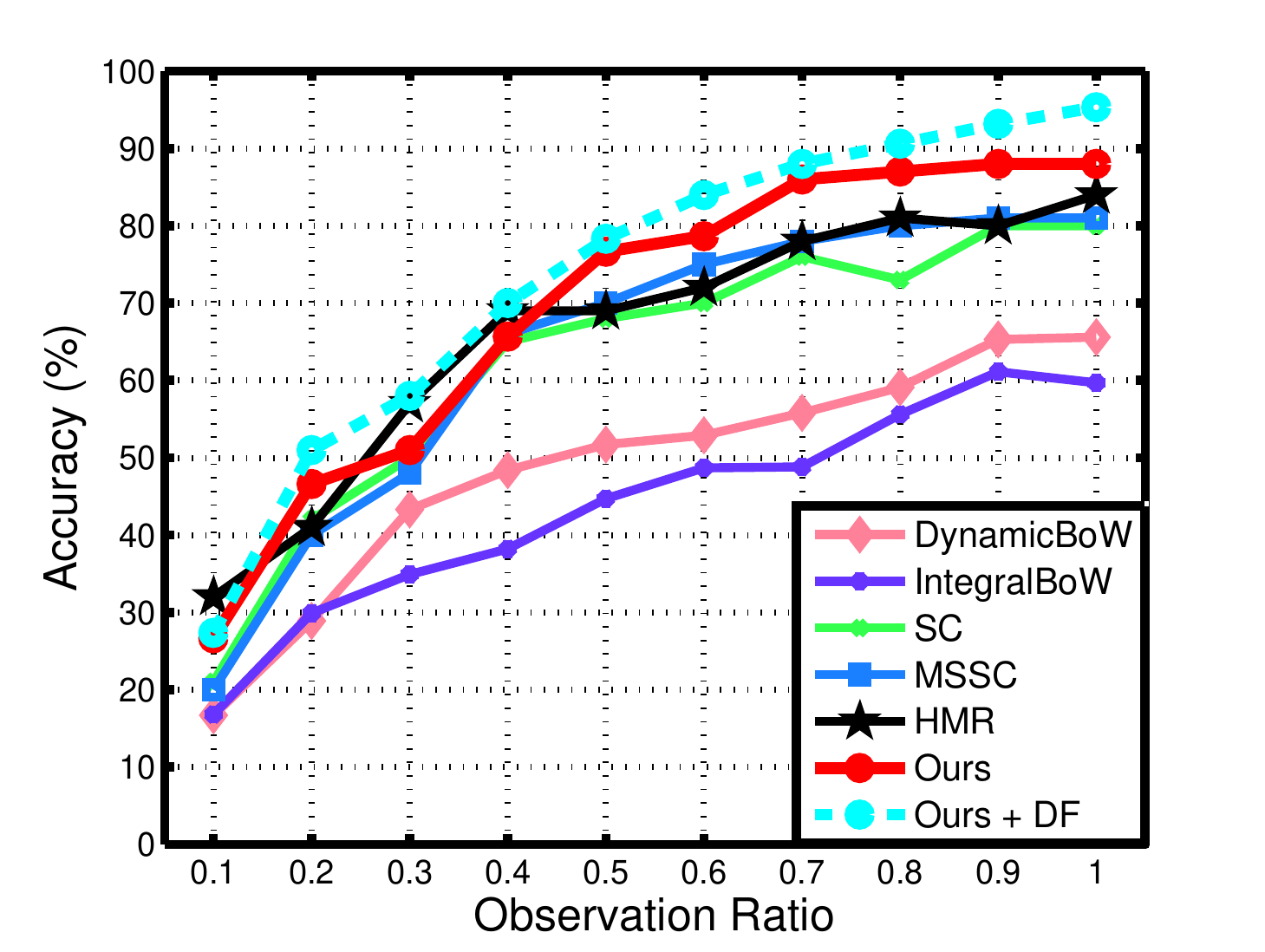}
			\label{fig:UT2_Prediction}\hspace{-0.25in}
			\includegraphics[height=1.4in]{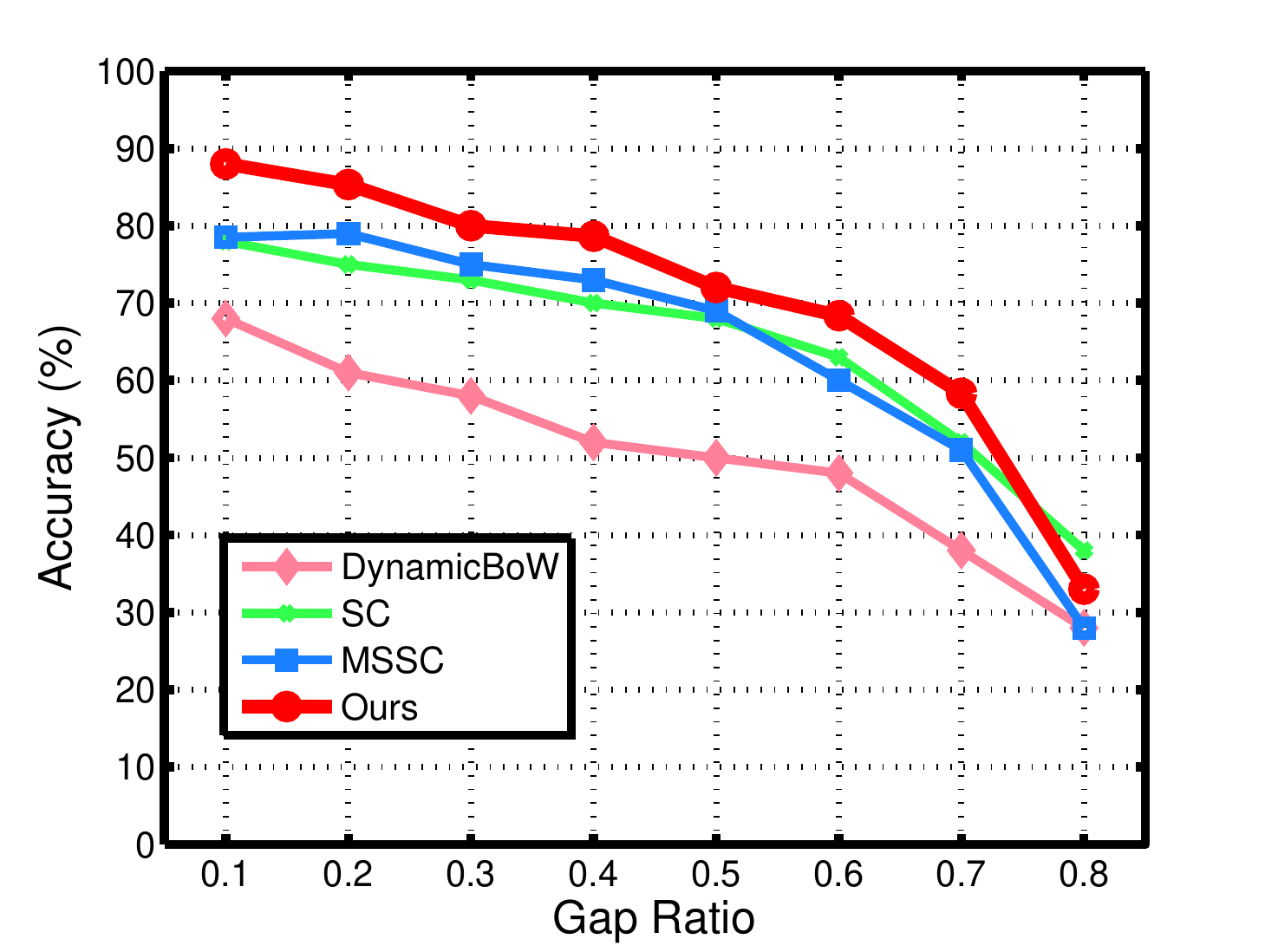}
			\label{fig:UT_gap1}\hspace{-0.25in}
			\includegraphics[height=1.4in]{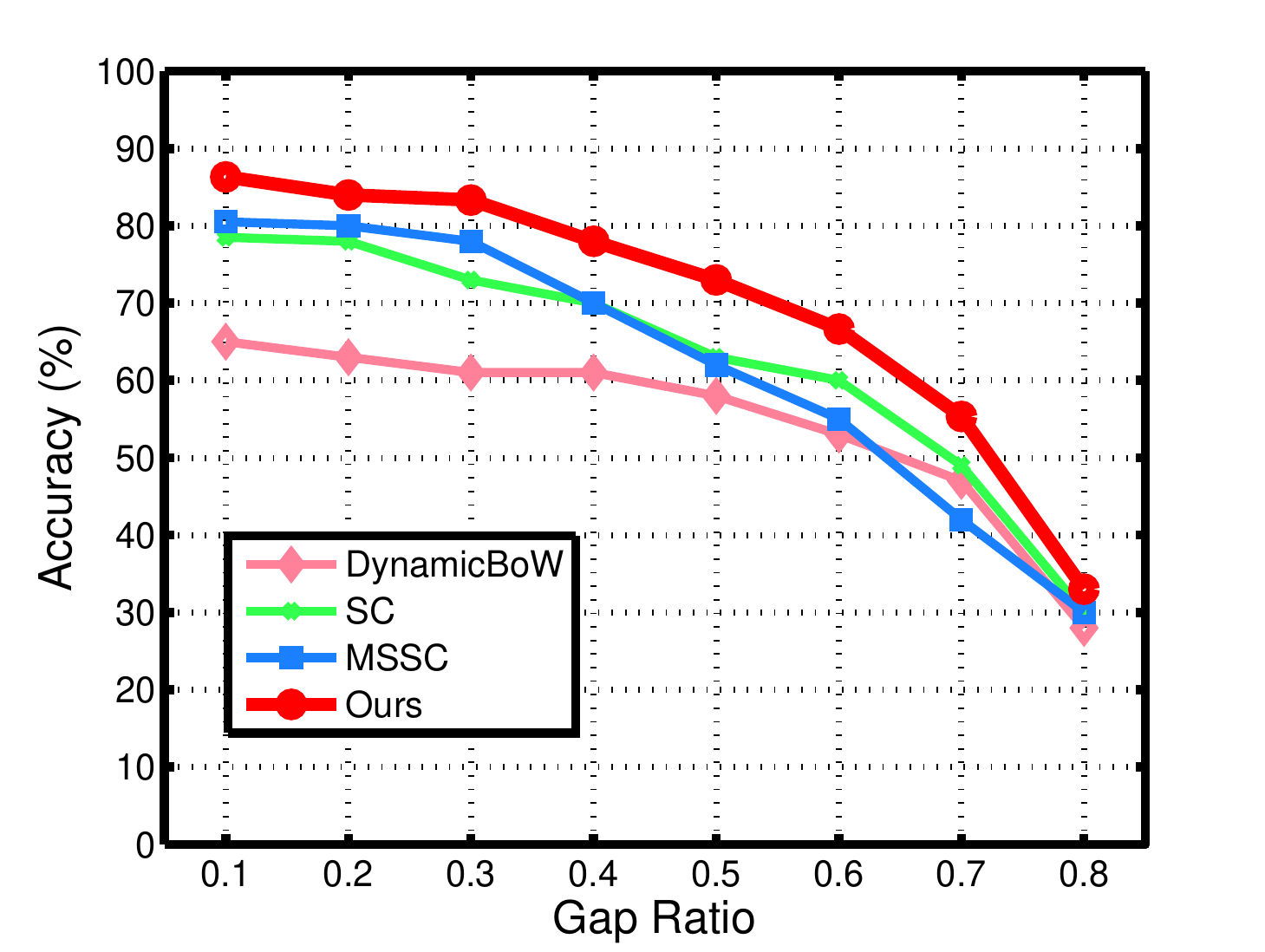}
			\label{fig:UT_gap2}\hspace{-0.25in}		
	\end{center}
	\label{fig:UT_outlier_known}
	\caption{Results of {\em early prediction} with {\em known} outlier frames on (a) UT-Interaction $\#1$ and (b) UT-Interaction $\#2$; Results of {\em gapfilling} with known outlier locations on (c) UT-Interaction $\#1$ and (d) UT-Interaction $\#2$.}
\end{figure*}

\begin{figure}[]
	\begin{center}
		\includegraphics[height=2.1in]{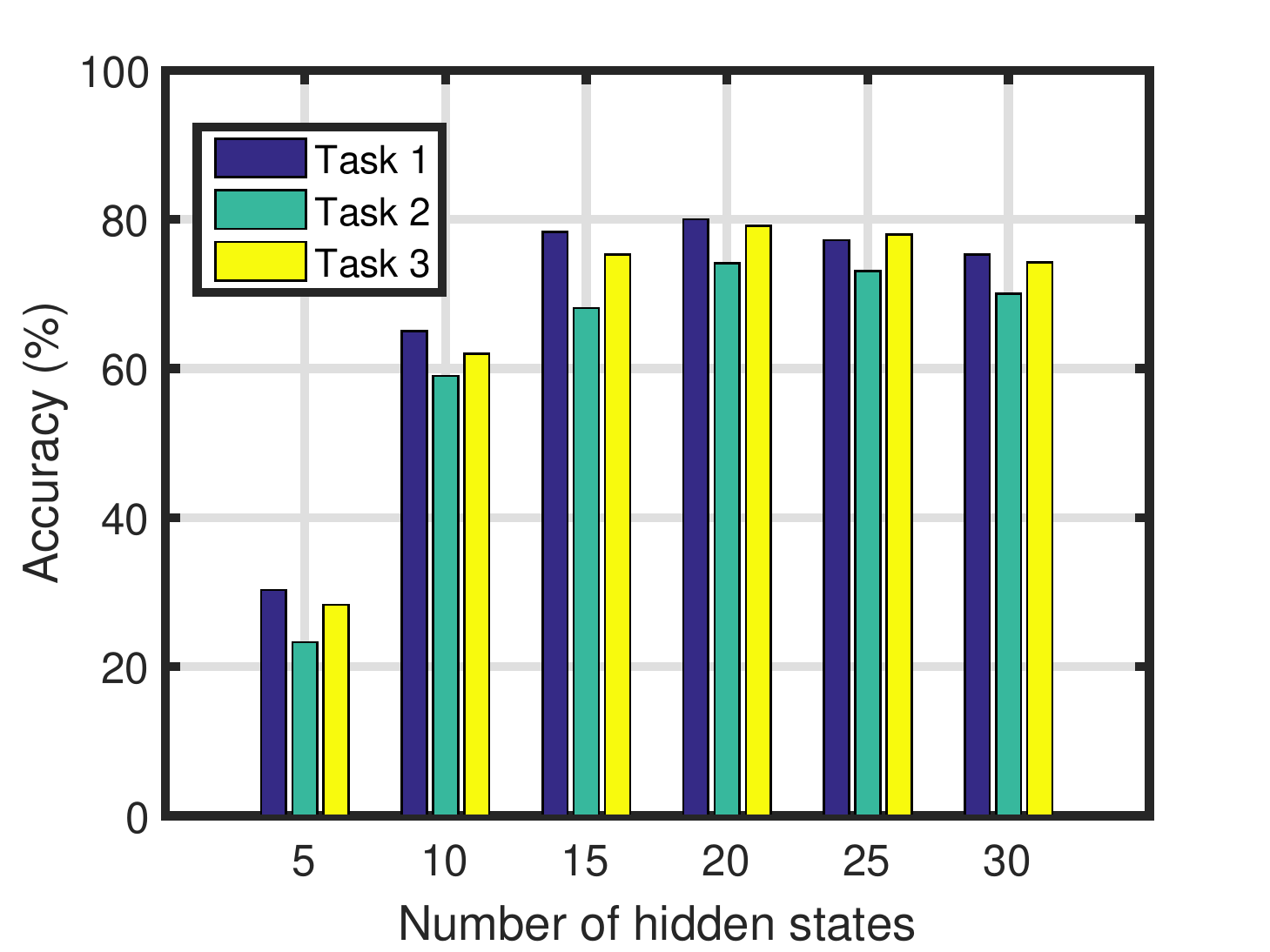}
	\end{center}
	\caption{The performance of our approach with different numbers of hidden states in HCRFs on all the three tasks.}
	\label{fig:hiddenpose}
\end{figure}


\subsection{Results on UT-Interaction Dataset}

Actions in the UT-Interaction dataset contains synthetic outlier frames and with {\em outlier ratios} from $0$ to $0.8$. Two sets of experiments are conducted. In the first one, approaches perform partially observed action recognition (POAR) in the case where the locations of outlier frames are {\em known} in advance. In this case, our approach augments only outlier segments with alternatives. We focus on comparing our approach to those working on actions with incomplete observation, and checking its advantage of borrowing alternatives from training data. In the second set, the locations of outlier frames are {\em unknown}. We focus on verifying whether our approach can detect outlier frames and pick proper alternatives to them, and result in remarkable performance gains over competing methods.

Our approach is compared with the same competing methods adopted in the experiments on the dataset we collected, except the method by Gowayyed~\etal~\cite{gowayyed13_}, which is designed on $3$D skeleton features and is not applicable on the UT-Interaction dataset. As mentioned, all the methods evaluated on this dataset adopt the bag-of-words model based on the cuboid descriptor.

\subsubsection{POAR with Known Outlier Locations}

We choose the setting of {\em gapfilling}~\cite{cao13_} and {\em early  prediction}~\cite{ryoo11_}, where outliers are the missing frames with known locations.
The former involves recognizing actions where the outlier (missing here) frames locate in the middle and thus, the observed frames are separated into two observed segments. The latter involves recognizing actions with missing frames at the end of the sequences.

The task of gapfilling is addressed under the assumption that the gap's location and duration are given. We select the state-of-the-art approaches, including {\em DynamicBoW}~\cite{ryoo11_}, {\em sparse coding} based method (SC)~\cite{cao13_}, {\em mixture of segments sparse coding} (MSSC)~\cite{cao13_}, for comparison.
Figure 8(a) and 8(b) report the performance of gapfilling by all the evaluated approaches on UT-Interaction datasets $\#1$ and $\#2$, respectively. Our approach is consistently superior to all other methods under different outlier ratios. We think the reason is that the compared approaches simply neglect the missing part, while our approach borrows extra alternatives to enrich the information for prediction, and connects the whole action for further temporal regularization.

We choose some of the state-of-the-art methods for comparison in early prediction, including {\em IntegrateBoWs}~\cite{ryoo11_}, {\em DynamicBoW}~\cite{ryoo11_}, {\em sparse coding} based method (SC)~\cite{cao13_}, {\em mixture of segments sparse coding} (MSSC)~\cite{cao13_}, {\em pose filter based hidden random conditional fields} (PF-HCRFs)~\cite{banerjee14_}, and {\em hierarchical movemes representation} (HMR)~\cite{lan14_}. To evaluate the performance of feature representation leaned by CNNs-based methods to action recognition, we also adopted the deep learning-based features (DF) extracted by using the two-stream architecture in~\cite{feichtenhofer16_}. The resultant method is denoted by {\em Ours+DF}.

\begin{figure*}[t]
		\hspace{-0.085in}
		\includegraphics[height=1.3in]{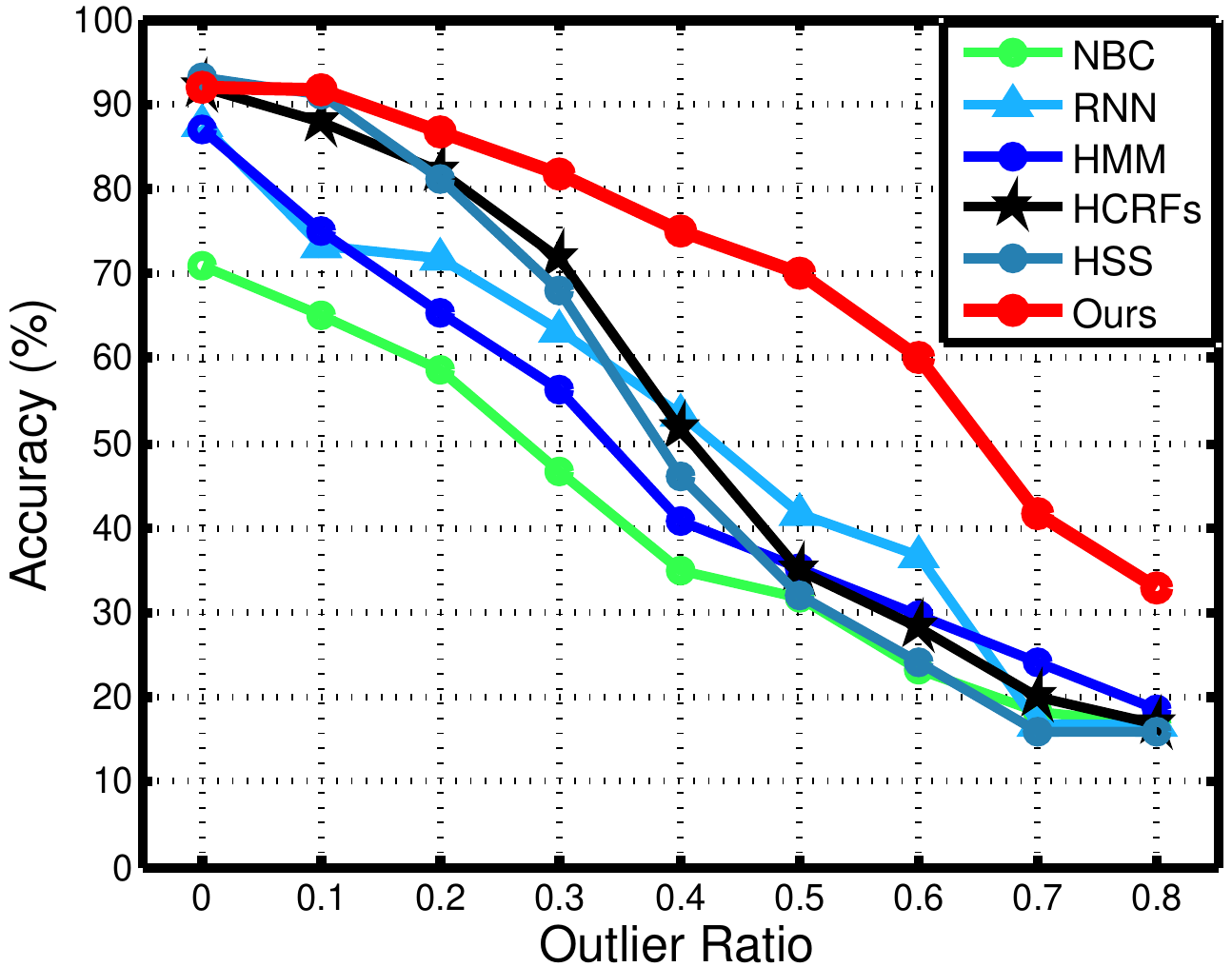}
			\label{fig:OT1_gapfilling_noise_ut1}\hspace{-0.05in}
		\includegraphics[height=1.3in]{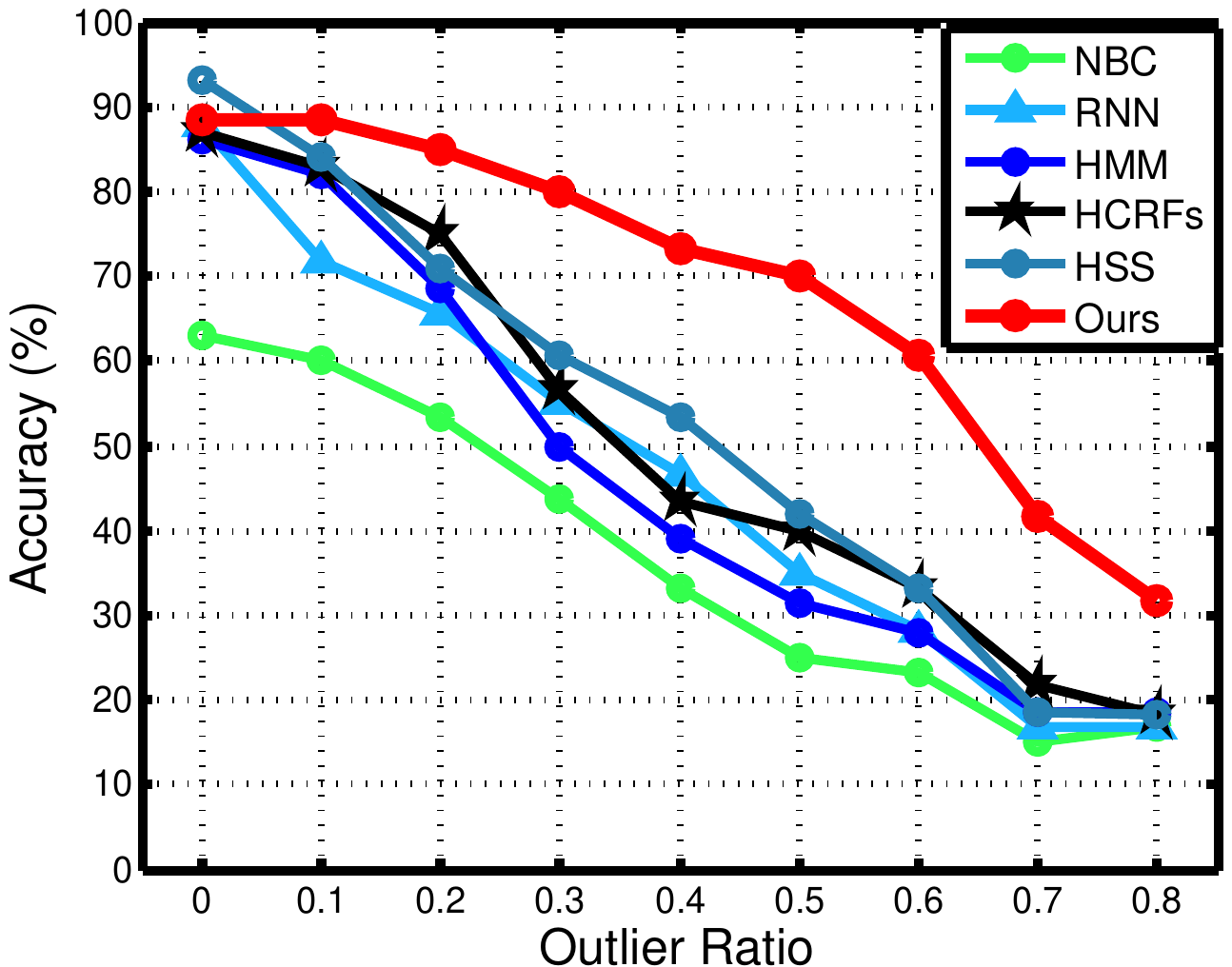}
			\label{fig:OT1_gapfilling_noise_ut2}\hspace{-0.05in}
 		\includegraphics[height=1.3in]{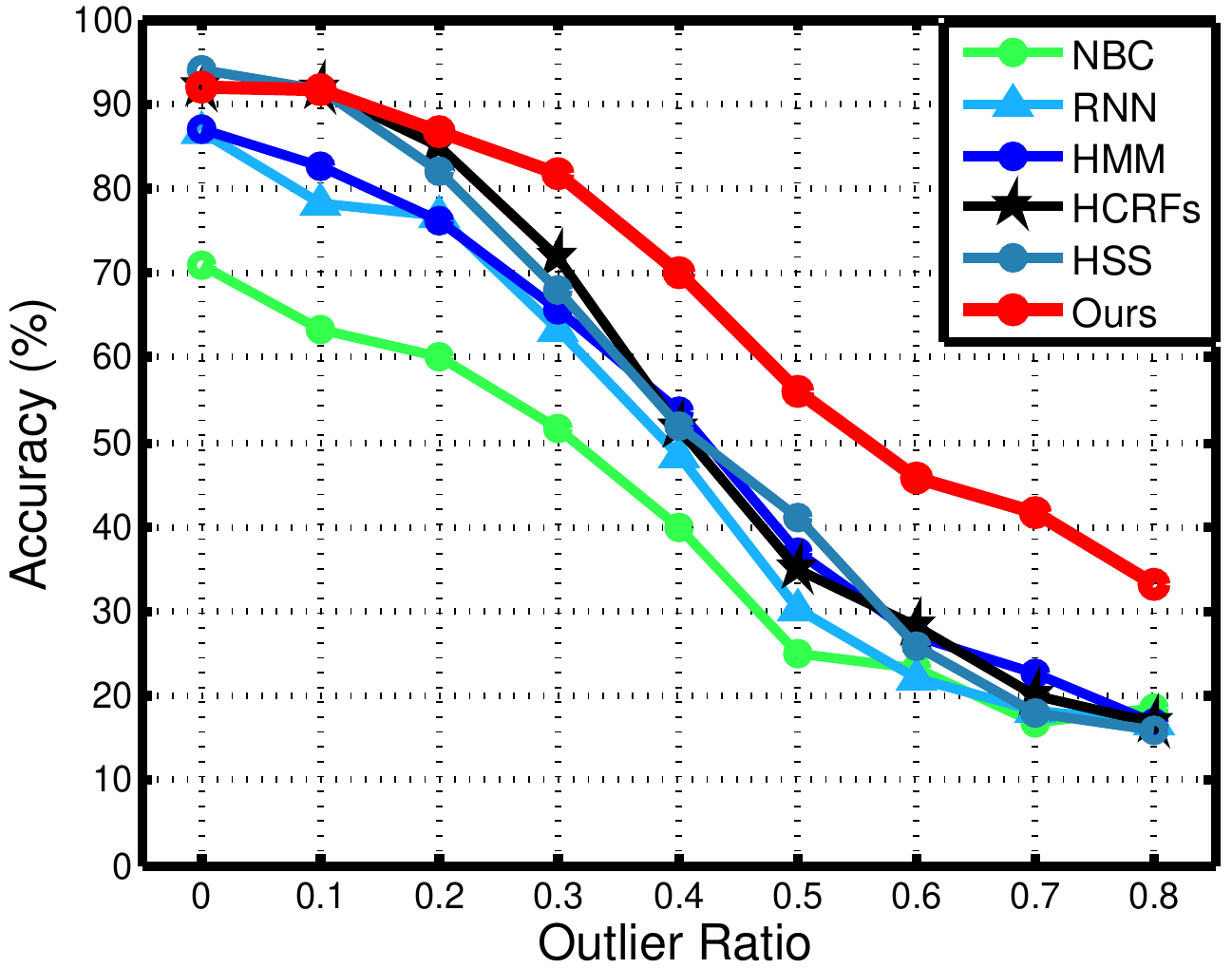}
			\label{fig:OT2_random_noise_ut1}\hspace{-0.05in}
		\includegraphics[height=1.3in]{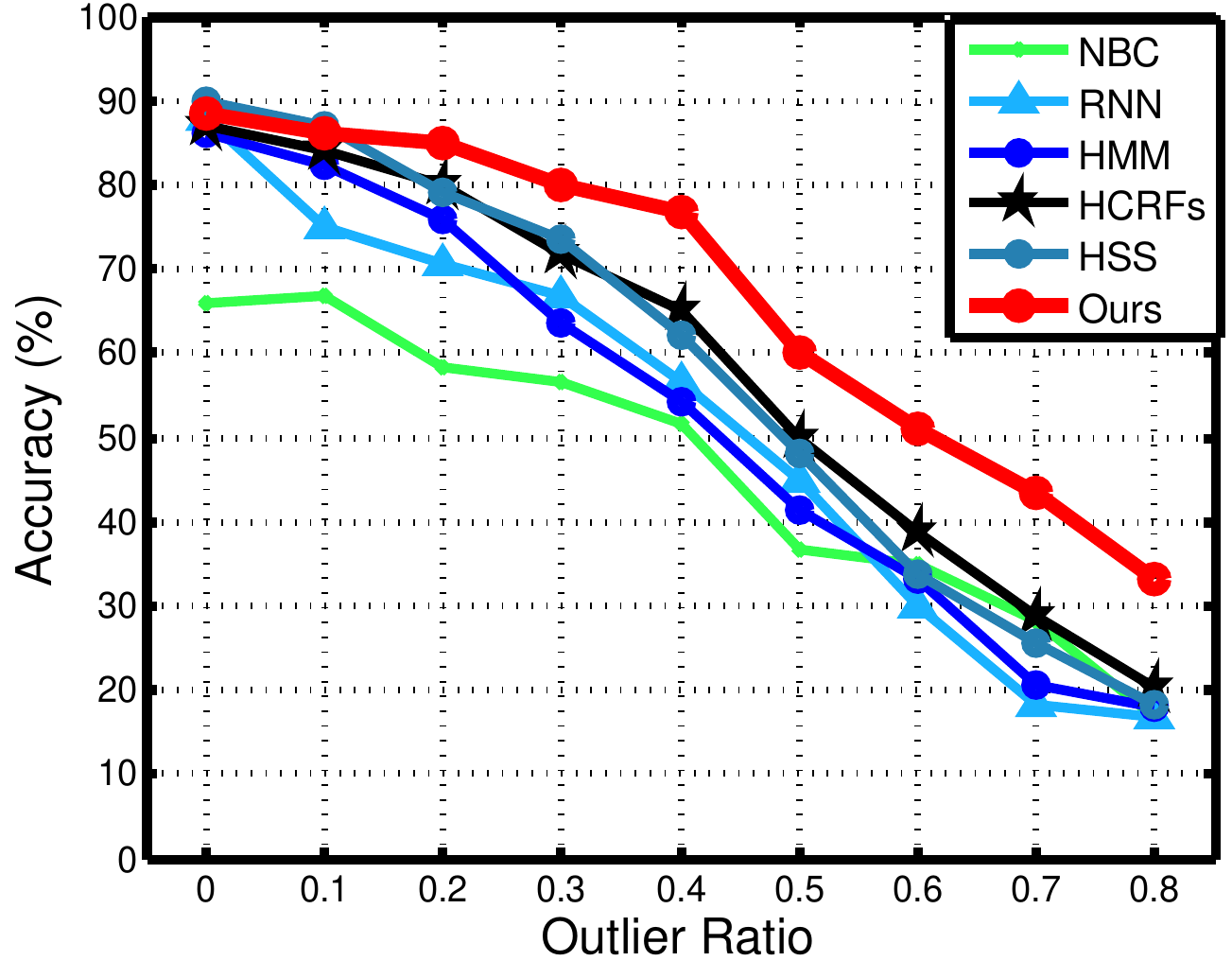}
			\label{fig:OT2_random_noise_ut2}\hspace{-0.05in}
	\caption{Results of POAR with {\em unknown} outlier locations. Gapfilling on (a) UT-Interaction $\#1$ and (b) UT-Interaction $\#2$. Randomly located outliers on (c) UT-Interaction $\#1$ and (d) UT-Interaction $\#2$.}
	\label{fig:OT}
\end{figure*}

\begin{figure}[t]
	\begin{center}
			\hspace{-0.08in}
			 \includegraphics[height=1.25in]{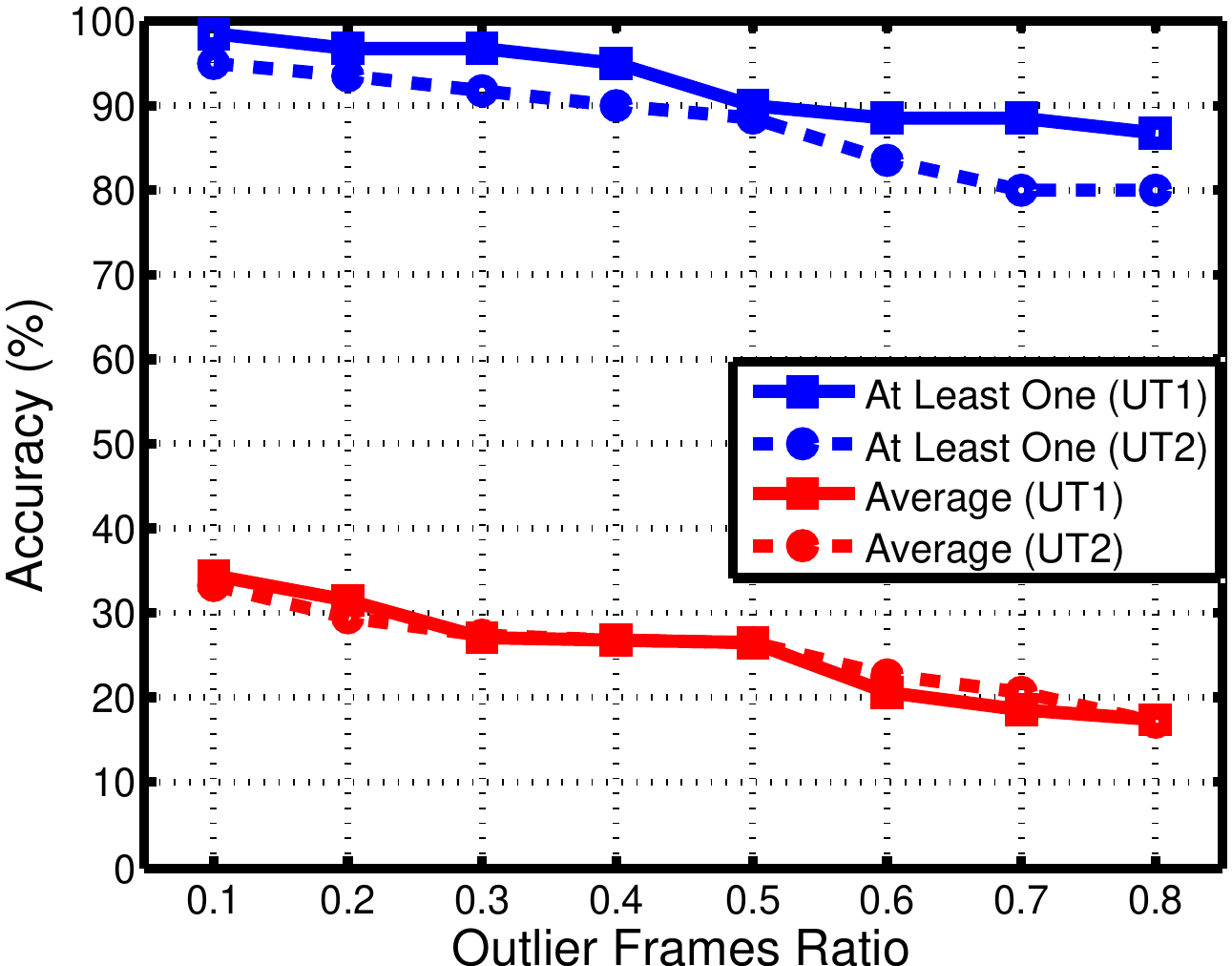}
			\label{fig:UTR_correctness}
			\hspace{-0.08in}
			\includegraphics[height=1.3in]{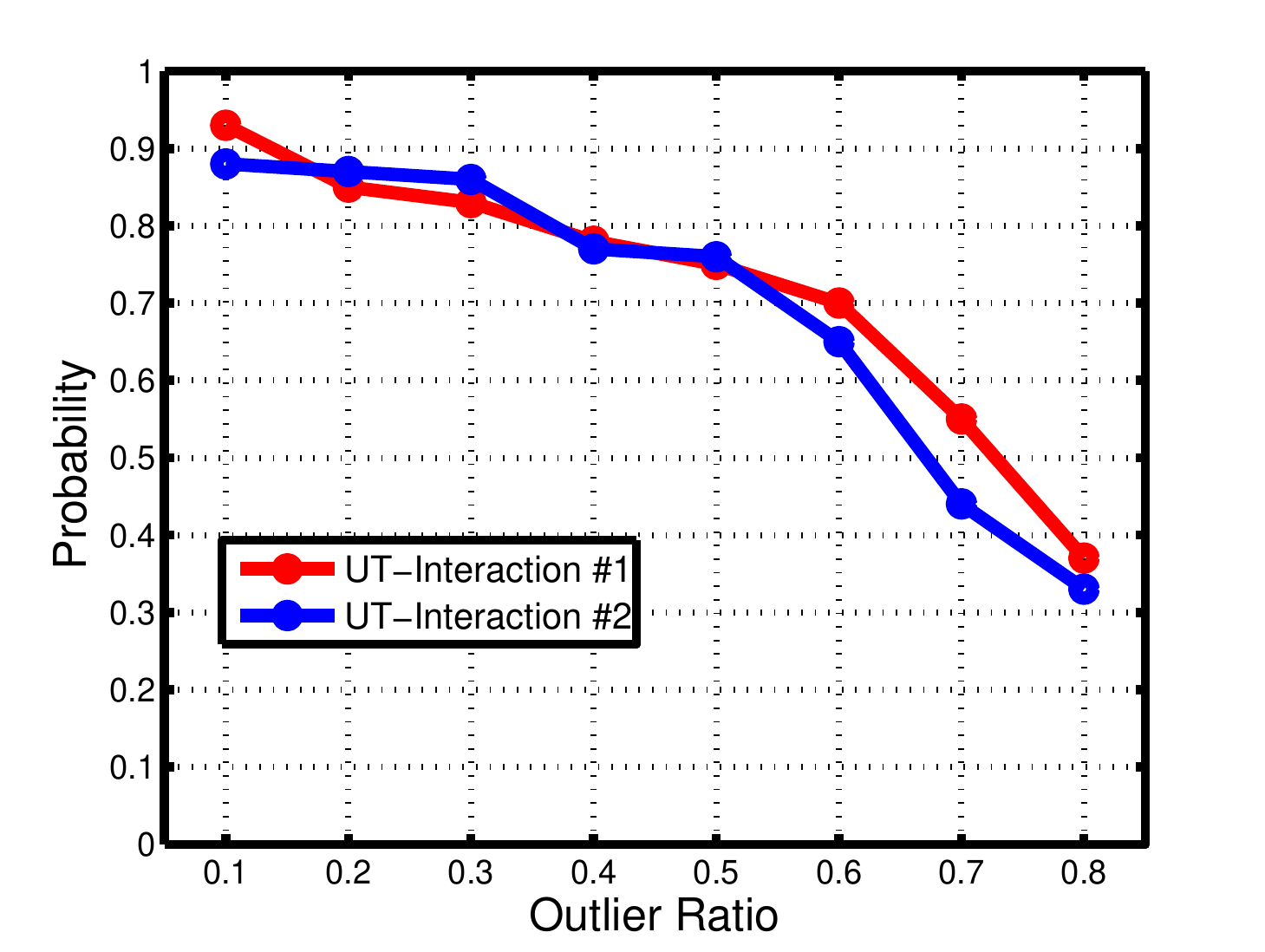}
			\label{fig:correct_outlier_replacement}
	\end{center}
    \vspace{-0.1in}
	\caption{(a) Alternative quality and (b) probabilities of correct outlier replacement under various outlier ratios. See the text for the details.}
	\label{fig:UT_correctness}
\end{figure}

Figure 8(a) and Figure 8(b) show the performance of the competing approaches and our approach on the UT-Interaction dataset $\#1$ and $\#2$, respectively. The recognition rates of each approach with different fractions of the observed segments in videos, \ie,~{\em observation ratios}, are given. Figure~8(a) shows that HCRFs-based approaches, such as our approach and PF-HCRFs~\cite{banerjee14_}, perform better than sparse coding based methods, \eg~SC and MSSC, and bag-of-words approaches, \eg~DynamicBow and IntegralBow. The main reason is that HCRFs employ hidden states in a chain structure in its representation, and the implicit temporal coherence in videos is better modeled in the latent space. Actions in UT-Interaction dataset $\#2$ are noisier than those in $\#1$. It can be seen in Figure 8(b) that the sparse coding based methods, SC and MSSC, are robust to noises and achieve comparable performance to HMR~\cite{lan14_} on UT-Interaction $\#2$. However, since the likelihood at each action segment is estimated independently, SC or MSSC would neglect temporal coherence among the observed parts. Our approach based on HCRFs employs temporal coherence information of the observed parts, and performs favorably against SC and MSSC.
Our approach with deep learning-based features (DF) performs slightly better than with the ordinary cuboid descriptor-based BoWs features in the both two datasets. The performance gain of adopting features learned by deep neural networks is not evident in our cases. The reason is that temporal convolutions are employed so that outlier frames make the computed features at their temporally nearby locations corrupt.

Compared to PF-HCRFs and HMR, our method achieves superior or similar performance (though is worse sometimes) as shown in Figures 8(a) and 8(b). We owe this to the reason that the unobserved part can be replaced by alternatives borrowed from training data in our approach, so carries the time-varying information. Then, by using both the observed part and the borrowed alternatives, temporal regularization is attainable to facilitate recognition in our approach.

The results in Figure~8 demonstrate that our approach can achieve favorable performance in comparison to the state-of-the-art approaches in POAR with known outlier frames. More importantly, our approach can carry out POAR even when the locations of outlier frames are unknown, as shown in the following. This property distinguishes our approach from the approaches compared in both gapfilling and early prediction.

\subsubsection{POAR with Unknown Outlier Locations}

Two settings are adopted for POAR with unknown outlier locations. The first one is still gapfilling, but the locations of outliers are assumed unknown. The second setting involves randomly located outlier frames whose locations in actions are arbitrarily generated. As in the self-collected dataset, our approach is compared with NBC, RNNs~\cite{martens11_}, HMM, HCRFs~\cite{quattoni07_}, and HSS~\cite{song13_} in this dataset. All approaches adopt the same bag-of-words representation by using the cuboid descriptor.

Figure~\ref{fig:OT} reports the performance of all evaluated approaches in the two settings on the UT-Interaction dataset $\#1$ and $\#2$. Except NBC, all methods achieve similar performance when no outlier presents. As the outlier ratio increases, our approach is significantly better than any competing approach. When the ratio is $0.5$, our approach achieves at least $30\%$ higher accuracy rates than any competing approaches in gapfilling and at least $10\%$ higher in the setting of using randomly located outliers. The results confirm the effectiveness of our approach in outlier detection and handling. Comparing the results in Figure~8 and Figure~\ref{fig:OT}, it can be observed that the performance gains of using our approach are more remarkable with the unknown locations of the outlier frames than with the known locations. This is because our approach can integrate outlier detection and alternative selection into prediction. To the best of our knowledge, this nice property distinguishes our approach from all the existing approaches.

\subsubsection{Alternative Quality Analysis}

To gain insight into why our method works well on POAR, we first analyze the quality of alternative segments borrowed from training data. We consider an alternative is {\em accurate} if it and the original segment belong to actions of the same class. For each augmented segment, we compute the average accuracy of its alternatives. We also measure the probability that at least one of its alternatives is accurate. Figure~11(a) show the two statistics under different outlier ratios. The results show that our alterative augmentation works well. With a high probability there exists at least one accurate alternative to a segment.

Moreover, we compute the probability that an outlier segment is replaced by an accurate alternative by our approach. As mentioned previously, the alternative with the maximal value in the potential function is selected to replace the original segment. The selected alternative is considered correct if it is from the training action of the same category. The {\em probability of correct replacement} under various outlier ratios are reported in Figure~11(b). It can be observed that more than $75\%$ outliers are correctly replaced when the outlier ratio is not higher than $0.5$. It reveals the main reason why our approach still works well when outliers occur.

\section{Conclusions}
We have introduced an approach to recognizing partially observed actions. We leverage the mutual dependency between video segments, and augment each segment of an action with extra alternatives borrowed from training data. When working on the augmented actions, our approach integrates {\em outlier segment detection} and {\em alternative selection} into the process of action recognition. To the best of our knowledge, such a generalization of action recognition is novel. Our approach is comprehensively evaluated on two datasets. It works with different features, recognizes actions with either synthetic or real outliers, and accomplishes gapfilling, full- and partially-observed action recognition. Experimental results demonstrates its effectiveness. For future study, we will aim to extend this approach to handle not only segment-level but also region- or trajectory-level outliers for advanced spatiotemporal analysis and further performance enhancement.


\bibliographystyle{ieee}
\bibliography{HAR}

\end{document}